\documentclass[conference]{IEEEtran}
\IEEEoverridecommandlockouts

\usepackage{cite}
\usepackage{amsmath,amssymb,amsfonts}
\usepackage{algorithmic}
\usepackage{graphicx}
\usepackage{textcomp}
\usepackage{xcolor}
\usepackage{etoolbox}
\usepackage[normalem]{ulem}
\usepackage{multirow}
\usepackage{caption}
\usepackage[section]{placeins}
\usepackage{float}
\usepackage{bbding}
\usepackage{color}
\usepackage{hyperref}
\usepackage[normalem]{ulem}
\usepackage{tabularx}
\usepackage{multirow}
\usepackage{amsmath}

\renewcommand{\baselinestretch}{1.0}

\hypersetup{
    colorlinks=true,
    linkcolor=blue,
    filecolor=blue,      
    urlcolor=blue,
    citecolor=black,
}

\def\BibTeX{{\rm B\kern-.05em{\sc i\kern-.025em b}\kern-.08em
    T\kern-.1667em\lower.7ex\hbox{E}\kern-.125emX}}
\begin{document}

\makeatletter
\patchcmd{\@makecaption}
  {\scshape}
  {}
  {}
  {}
\makeatletter
\patchcmd{\@makecaption}
  {\\}
  {:\ }
  {}
  {}
\makeatother

\title{MPCGNet: A Multiscale Feature Extraction and Progressive Feature Aggregation  Network Using Coupling Gates for Polyp Segmentation\\
}
\author{\IEEEauthorblockN{1\textsuperscript{st} Wei Wang}
\IEEEauthorblockA{\textit{School of Computer and} \\ 
\textit{Communication Engineering,} \\
\textit{Changsha University of Science}\\
\textit{and Technology}\\
Changsha, China \\
wangwei@csust.edu.cn}
\and
\IEEEauthorblockN{2\textsuperscript{nd} Feng Jiang}
\IEEEauthorblockA{\textit{School of Computer and} \\ 
\textit{Communication Engineering,} \\
\textit{Changsha University of Science}\\
\textit{and Technology}\\
Changsha, China \\
318523296@qq.com}
\and
\IEEEauthorblockN{3\textsuperscript{rd} Xin Wang\textsuperscript{*}}
\IEEEauthorblockA{\textit{School of Computer and} \\ 
\textit{Communication Engineering,} \\
\textit{Changsha University of Science}\\
\textit{and Technology}\\
Changsha, China \\
wangxin@csust.edu.cn}}

\maketitle

\begin{abstract}
Automatic segmentation methods of polyps is crucial for assisting doctors in colorectal polyp screening and cancer diagnosis. Despite the progress made by existing methods, polyp segmentation faces several challenges: (1) small-sized polyps are prone to being missed during identification, (2) the boundaries between polyps and the surrounding environment are often ambiguous, (3) noise in colonoscopy images, caused by uneven lighting and other factors, affects segmentation results. To address these challenges, this paper introduces coupling gates as components in specific modules to filter noise and perform feature importance selection. Three modules are proposed: the coupling gates multiscale feature extraction (CGMFE) module, which effectively extracts local features and suppresses noise; the windows cross attention (WCAD) decoder module, which restores details after capturing the precise location of polyps; and the decoder feature aggregation (DFA) module, which progressively aggregates features, further extracts them, and performs feature importance selection to reduce the loss of small-sized polyps. Experimental results demonstrate that MPCGNet outperforms recent networks, with mDice scores 2.20\% and 0.68\% higher than the second-best network on the ETIS-LaribPolypDB and CVC-ColonDB datasets, respectively. 
\end{abstract}


\section{Introduction}
Colorectal cancer (CRC) ranks as the second leading cause of cancer-related deaths and stands as the third most frequently diagnosed malignancy on a global scale~\cite{b17}. Colorectal polyps, common pathological growths in the digestive system, are closely linked to the pathogenesis of CRC and are generally regarded as precancerous lesions in their development. Colonoscopy is a widely utilized technique for screening colorectal cancer~\cite{b32}. However, due to the substantial variability in the appearance of colorectal polyps during colonoscopy, traditional manual examination methods are highly dependent on the operator's subjective experience, which can result in missed diagnoses and misdiagnoses. Consequently, developing automated polyp segmentation methods based on computer vision has become a critical research focus in medical imaging analysis.
\begin{figure}
\centering
\includegraphics[width=0.5\textwidth]{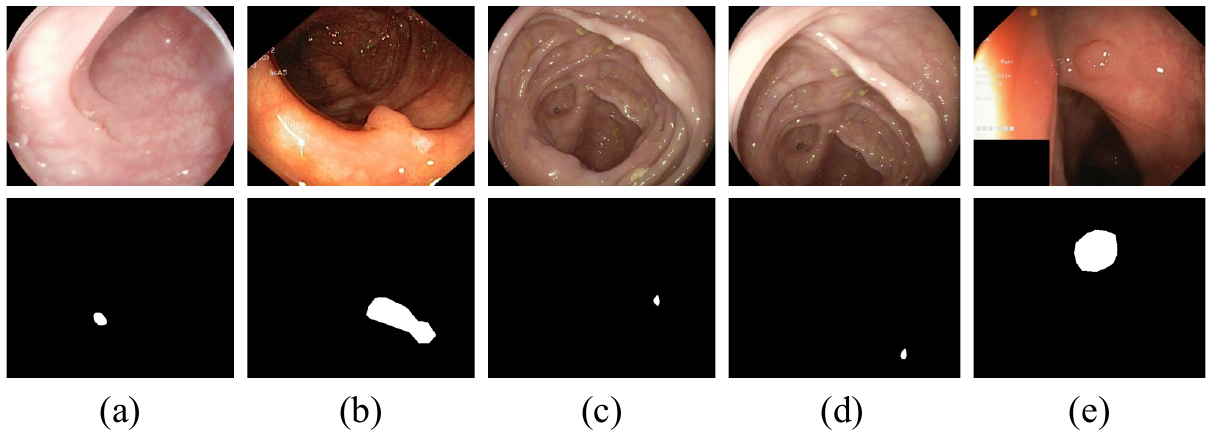}
\caption{Some examples of polyp images.}
\label{example}
\end{figure}

To design an effective method for polyp segmentation, it is essential to observe the characteristics of polyp images and identify the associated challenges. As illustrated in Fig.~\ref{example}, several issues can be observed. In Fig.~\ref{example}(a), the left edge of the polyp exhibits blurred boundaries. Fig.~\ref{example}(b)(e) show images with uneven lighting, leading to inconsistent brightness and the introduction of noise. Additionally, Fig.~\ref{example}(c)(d) depict small-sized polyps, which are prone to being overlooked during the segmentation process.
To address the aforementioned challenges, this paper proposes a multiscale feature extraction and progressive feature aggregation network using coupling gates for polyp segmentation (MPCGNet). First, the Coupling Gates Multiscale Feature Extraction (CGMFE) module is introduced to effectively extract local features, such as boundaries, while suppressing noise caused by factors such as uneven lighting through the use of coupling gates. The Windows Cross Attention Decoder (WCAD) module is designed to restore boundary details and other fine features after identifying the precise location of the polyp. The Decoder Feature Aggregation (DFA) module is utilized to capture small-sized polyps by progressively expanding the scope of captured features and re-extracting aggregated features from multiple perspectives. The coupling gates are then used to perform feature importance selection and prevent small polyps from being overlooked. The main contributions of this study are outlined as follows: (1) A novel polyp segmentation network, MPCGNet, is proposed; (2) Coupling gates are introduced to filter noise and perform feature importance selection; (3) Three key modules---CGMFE, WCAD, and DFA---are designed to address the three major challenges in polyp segmentation; (4) The proposed network outperforms recent polyp segmentation networks.

\section{Related Work}
% \section{RELATED WORK}
\subsection{Polyp Segmentation}\label{AA}
\subsubsection{General-Purpose Methods}
% With the development of deep learning, particularly the advent of Convolutional Neural Networks (CNNs) capable of extracting local features, a new approach for medical image segmentation has emerged. FCN~\cite{b20} was the first to introduce end-to-end pixel-level segmentation, laying the foundation for subsequent medical image segmentation methods. UNet~\cite{b21} proposed a classic encoder-decoder architecture that integrates high-level and low-level features through skip connections, becoming a benchmark model for medical image segmentation. UNet++~\cite{b22} further optimized the skip connection pathways to mitigate the semantic gap between multi-resolution features. With the rise of Transformers, TransUNet~\cite{b23} embedded Transformer modules into the UNet framework, enhancing the ability to capture long-range dependencies. Swin-Unet~\cite{b24} leveraged the hierarchical Swin Transformer to further improve multiscale feature modeling capabilities, setting a new standard for medical image segmentation.
General-purpose methods leverage the flexibility and scalability of deep learning models. The classic U-Net architecture~\cite{b21}, with its encoder-decoder structure, is widely regarded as a cornerstone for medical image segmentation tasks. Building on this cornerstone, advanced convolutional neural networks (CNNs), such as ResNet~\cite{b34} and DenseNet~\cite{b35}, have been employed as encoders to improve feature extraction. Recently, Transformer-based models, such as Swin-UNet~\cite{b36}, have been introduced to medical segmentation, employing attention mechanisms to extract semantic information at both local and global levels. However, these general-purpose methods often struggle to capture the unique characteristics of colonoscopy images, leading to suboptimal segmentation accuracy and robustness.

\subsubsection{Task-Specific Methods}
Task-specific methods focus on the particular challenges of colon polyp segmentation including missed detection of small polyps, indistinct boundaries, and interference from noise. To address these issues, researchers have proposed various improved methods. In response to the issue of small-sized polyps being easily missed during identification, M3FPolypSegNet~\cite{b33} employs a multi-frequency feature fusion strategy, utilizing multi-frequency encoders and frequency-space pyramid pooling modules to effectively locate and segment small polyps. HSNet~\cite{b12} integrated a hybrid semantic network with a multi-scale prediction module to strengthen multi-scale modeling for small polyps. To address boundary blurriness, PraNet ~\cite{b11} enhanced boundary awareness through a parallel reverse attention mechanism. ECTransNet~\cite{b14} employed a multiscale edge complementary module to enhance boundary features, significantly improving segmentation accuracy. Similarly, MegaNet~\cite{b15} addressed the interference caused by the similarity between boundaries and the background through a multiscale edge-guided attention network. For uneven illumination and noise interference, CASCADE~\cite{b13} utilized a cascaded attention decoding module to progressively eliminate noise features layer by layer, thereby enhancing segmentation robustness. LHONet~\cite{b16} mimick`ed the human hierarchical observation pattern, enhancing segmentation from coarse to fine, effectively mitigating the effects of noise and uneven illumination. LSSNet~\cite{b27} suppressed noise information and reinforced feature expression in blurred regions by leveraging local feature augmentation and shallow feature supplementation structures, employing a multiscale feature extraction module and a shallow feature supplementation module.
\subsection{Gating Mechanism}
Hochreiter et al.~\cite{b25} proposed the Long Short-Term Memory (LSTM) network, which effectively controls information flow through forget, input, and output gates, addressing the gradient vanishing problem of traditional recurrent neural networks when handling long sequences. Gers et al.~\cite{b26} extended LSTM by introducing a structure with forget gates, enhancing its ability to capture long-term dependencies. This structure allows the model to dynamically adjust the forget gate parameters, enabling more flexible retention or discarding of historical information. Cho et al.~\cite{b18} utilized gating mechanisms in sequence modeling to dynamically select important features; the reset gate adjusts the influence of the previous time step's hidden state on the current candidate state, while the update gate modulates the integration ratio between current input and historical hidden states. Dauphin et al.~\cite{b19} proposed a simplified gating mechanism that divides input features into two parts: one for output computation and the other for generating gating signals via a sigmoid function. This mechanism controls the retention or suppression of the first part's features, effectively filtering out irrelevant information. Wang et al.~\cite{b38} introduced a Gated Convolutional Neural Network (GCN) for semantic segmentation, where entropy maps guide a gating mechanism to adjust the importance of feature maps from different layers. This allows the model to prioritize uncertain pixels and utilize finer details from lower layers, enhancing segmentation accuracy. Oktay et al.~\cite{b39} developed Attention U-Net, which utilizes attention gates to adaptively modify the weighting of various regions within the image. This mechanism emphasizes important areas and suppresses irrelevant regions, particularly improving segmentation accuracy for small objects in medical images.

\section{Methods}
\begin{figure*}[h]
\centering
\includegraphics[scale=1]{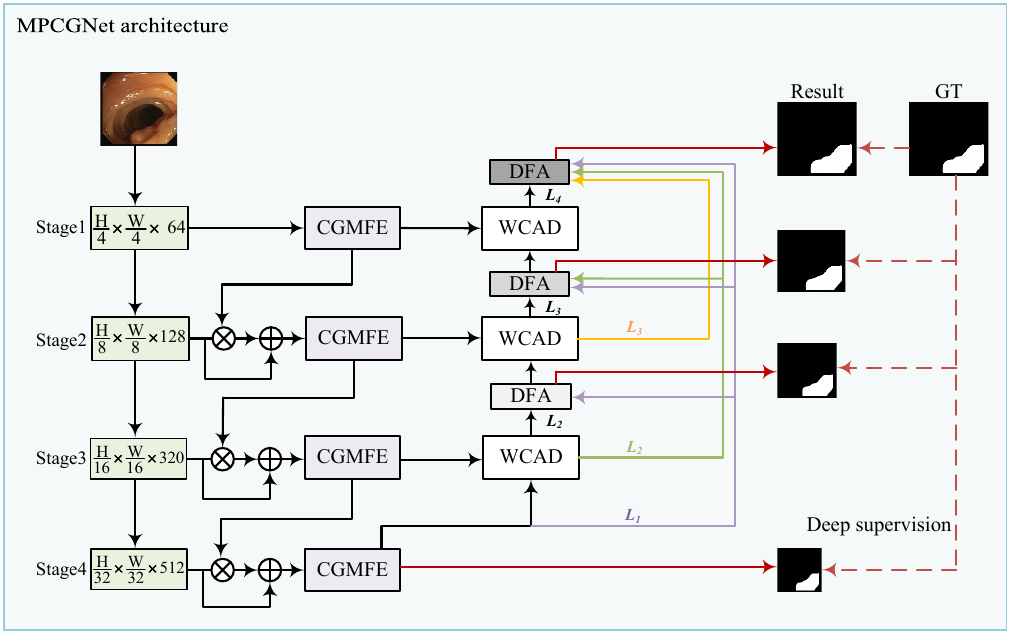}
\caption{The architecture of MPCGNet}
\label{architecture}
\end{figure*}
Fig.~\ref{architecture} presents an illustration of the architecture of our proposed method. A pretrained PVT v2~\cite{b2} serves as the backbone. The CGMFE module performs multiscale extraction of local features and suppresses noise using coupling gates. The encoder of each stage interacts with the CGMFE module of the previous stage through element-wise multiplication and skip connection, and the result is then fed into the CGMFE module of the current stage. The WCAD module employs a cross-attention mechanism to gradually fuse shallow low-level features with deep high-level features, enhancing the clarity of local details. The DFA module progressively aggregates decoding features across layers, performing feature extraction and feature importance selection in spatial, channel, and global dimensions. During training, a 4-stage deep supervision approach is employed. During testing, the final result is only generated from the top DFA module.
\begin{figure*}
\centering
\includegraphics[scale=1]{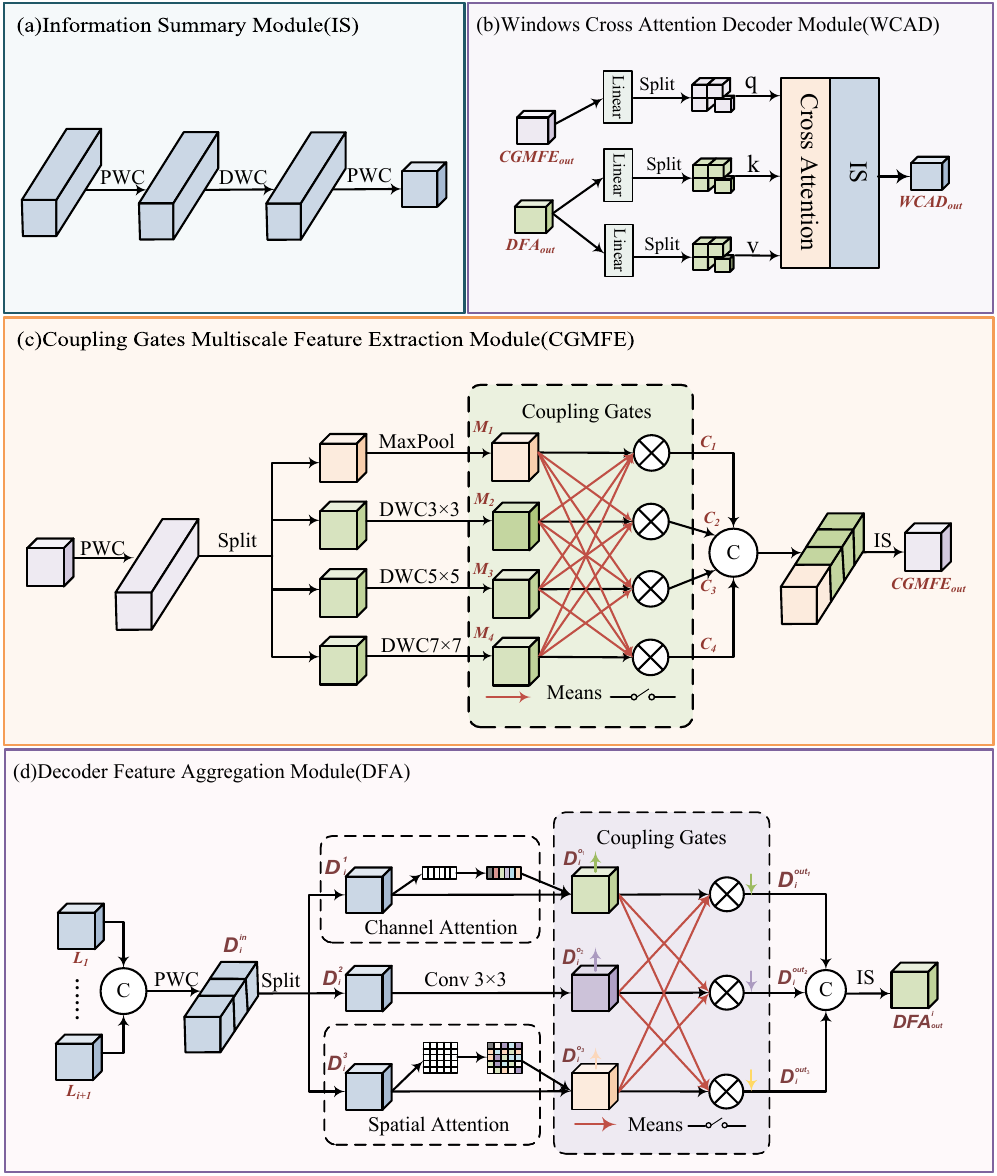}
\caption{(a) Structure of IS. (b) Structure of WCAD. (c) Structure of CGMFE. (d) Structure of DFA.}
\label{mynet}
\end{figure*}
\subsection{Information Summary Module}
The CGMFE and DFA modules proposed in this paper both adopt an inverted bottleneck structure\cite{b1}, leveraging the advantages of this design to map the feature space to higher dimensions, thereby enabling more comprehensive feature extraction. The information summary (IS) module is designed to assist subsequent modules in achieving the overall inverted bottleneck structure, as illustrated in Fig.~\ref{mynet}(a). First, a pointwise convolution is used to fuse channel features without altering the number of channels. Next, a depthwise convolution with a kernel size of 3 × 3 is employed for feature extraction. Finally, another pointwise convolution reduces the features to the corresponding number of channels. For an input feature map $x$, this process can be expressed as the formula:
\begin{equation}
{IS}_{out}=PWC\left({DWC}_{3\times3}\left(PWC\left(x\right)\right)\right)\
\end{equation}
where ${DWC}_{3\times3}$ is the depthwise convolution with kernel size of 3 × 3 and $PWC$ is the pointwise convolution. Similarly, the following text will use the same terminology.

\subsection{Coupling Gates Multiscale Feature Extraction Module}
To effectively extract local features, suppress noise, bridge the encoder and decoder, and reduce the semantic gap between them, the coupling gates multiscale feature extraction (CGMFE) module is proposed. As shown in Fig.~\ref{mynet}(c), the CGMFE module employs four parallel branches for multiscale extraction of local features, consisting of max pooling and depthwise convolutions with kernel sizes of 3 × 3, 5 × 5, and 7 × 7, resulting in $M_{1},M_{2},M_{3},M_{4}$. The strong coupling relation~\cite{b3} is typically established in dual-branch fusion scenarios. However, in the case of multi-branch parallelism, it is unclear how to establish optimal strong coupling relation: branches with significant noise should not form strong coupling relation, whereas branches with overlapping or complementary local features should. Coupling Gates are designed to address this problem. The values of the coupling gates are constrained between 0 and 1 using a sigmoid function. The values of the coupling gates are converted to binary (0 or 1) using a threshold of 0.5; values exceeding 0.5 are assigned 1, while values of 0.5 or less are assigned 0. Each coupling gate is initially at a critical threshold and closed (0). For the four parallel branches, each branch is connected to the other three through coupling gates, which determine whether branch fusion occurs. This process can be expressed as the formula:
\begin{equation}
 C_{i}=M_{i} \times \prod_{j \neq i}\left(M_{j}\right)^{{gate}_{j i}}
\end{equation}
where $i,j\in 1,2,3,4$, ${ gate }_{j i}=0 \text { or } 1$. If the value of ${ gate }_{j i}$ is 1, it means the gate from branch $j$ to branch $i$ is opened, establishing the strong coupling relation between them. Otherwise, this relation is not established because the exponential operation turns all elements of $M_{j}$ into 1. $C_{i}$ denotes the feature map generated for branch $i$ after using coupling gates. Similarly, the following text will use the same terminology.

The outputs from the four branches are subsequently concatenated. Finally, the concatenated result is passed through the IS module for further feature extraction, ensuring that the entire module maintains the inverted bottleneck structure. This process can be expressed as the formula:
\begin{equation}
{CGMFE}_{out}=IS\left({ Concat }\left(C_{1}, C_{2}, C_{3}, C_{4}\right)\right)
\end{equation}

\subsection{Windows Cross Attention Decoder Module}
As shown in Fig.~\ref{mynet}(b), the windows cross attention decoder module (WCAD) employs a cross-attention mechanism to fuse deep high-level features from DFA (or stage 4's CGMFE) with shallow low-level features from CGMFE. This fusion restores feature details after locating the position of the polyp. First, the feature map from DFA (or stage 4's CGMFE) is processed through weight matrices ${W}_{k}$ and ${W}_{v}$ to generate the ${Key}(K)$ and ${Value}(V)$, respectively, while the feature map from CGMFE is processed through the weight matrix ${W}_{q}$ to generate the ${Query}(Q)$. The attention mechanism utilizes Expanded Window Multi-Head Self-Attention (EW-MHSA)~\cite{b4}. $Q$, $K$, and $V$ are partitioned into windows for attention matrix computation, after which the result is reconstructed to its original dimensions. The IS module is used as a feed-forward layer to further extract detailed features, such as boundaries. The entire process can be expressed as the formula:
\begin{equation}
{WCAD}_{out}=IS\left({ EW - MHSA }\left(Q, K, V\right)\right)
\end{equation}

\subsection{Decoder Feature Aggregation Module}
To reduce the loss of small-sized colorectal polyps, the decoder feature aggregation (DFA) module performs large-scale cross-layer feature aggregation and capture. The DFA module adopts a progressive feature aggregation approach, where features from stage 4’s CGMFE and the WCAD, which is located deeper than the current DFA layer, are aggregated at each step. As the layers become shallower, the aggregated features gradually increase, and the scope of captured features expands, as illustrated in Fig.~\ref{architecture}, where the color of the DFA module progressively deepens. After aggregating the features, the feature maps undergo extraction across spatial, channel, and global dimensions, with coupling gates employed to effectively perform feature importance selection.
The overall architecture of the DFA module is depicted in Fig.~\ref{mynet}(d). First, the deep-layer decoder features $\left\{L_1,\ldots,L_{i+1}\right\}$ from stage 4's CGMFE and the WCAD which is located deeper than the current DFA layer,  are concatenated. A pointwise convolution is then applied to fuse the channel features and adjust the number of channels. This process can be expressed as the formula:
\begin{equation}
{D}_{in}^i=PWC\left({ Concat }\left(L_{1}, \ldots, L_{i+1}\right)\right)
\end{equation}

The feature map $D_{in}^i$ is then divided into three parallel branches in the channel dimension: $D_{i}^1$, $D_{i}^2$, and $D_{i}^3$. The $D_{i}^1$ branch employs a Channel Attention (ChAtt) Module~\cite{b5} to extract channel features, resulting in $D_{i}^{o_{1}}$. The $D_{i}^2$ branch uses a convolution with a kernel size of 3 × 3 to extract global features, yielding $D_{i}^{o_{2}}$. The $D_{i}^3$ branch utilizes a Spatial Attention (SpAtt) Module~\cite{b5} to extract spatial features, producing $D_{i}^{o_{3}}$. This process can be expressed as the formula:
% \begin{small}
% \begin{equation}
% {D}_{i}^{o_{1}}=CA\left(D_{i}^1\right),\ {D}_{i}^{o_{2}}=Conv_{3\times3}\left(D_{i}^2\right),\ {D}_{i}^{o_{3}}=SA\left(D_{i}^3\right)
% \end{equation}
% \end{small}
\begin{equation}
\begin{aligned}
&{D}_{i}^{o_{1}}=ChAtt\left(D_{i}^1\right),\\
&{D}_{i}^{o_{2}}=Conv_{3\times3}\left(D_{i}^2\right),\\
&{D}_{i}^{o_{3}}=SpAtt\left(D_{i}^3\right)
\end{aligned}
\end{equation}
where $Conv_{3 \times 3}$ represents a convolution operation with a kernel size of 3 × 3, $ChAtt(\cdot)$ denotes the Channel Attention module, and $SpAtt(\cdot)$ denotes the Spatial Attention module.

Finally, the feature maps from the three branches establish optimal strong coupling relation through coupling gates, performing feature importance selection. These selected feature maps are then added to the previous feature maps via skip connections. Subsequently, the feature maps from the three branches are concatenated and fed into the IS module to further inspect for any 
overlooked small-sized polyps. This process can be expressed as the formula:
% \begin{equation}
% D_{i}^{{out}_{m}}=D_{i}^{o_{m}} \times \prod\left(D_{i}^{o_{n}}\right)^{{gate}_{n m}}+D_{i}^{o_{m}}    
% \end{equation}
% \begin{equation}
% DFA_{out}^i=IS\left({ Concat }\left(D_{i}^{{out}_{1}}, D_{i}^{{out}_{2}}, D_{i}^{{out}_{3}}\right)\right)   
% \end{equation}
\begin{equation}
\begin{aligned}
&D_{i}^{{out}_{m}}=D_{i}^{o_{m}} \times \prod_{n \neq m}\left(D_{i}^{o_{n}}\right)^{{gate}_{n m}}+D_{i}^{o_{m}},\\
&DFA_{out}^i=IS\left({ Concat }\left(D_{i}^{{out}_{1}}, D_{i}^{{out}_{2}}, D_{i}^{{out}_{3}}\right)\right)   
\end{aligned}
\end{equation}
where $m,n\in 1,2,3$, ${ gate }_{n m}=0 \text { or } 1$, $DFA_{out}^i$ denotes the output feature map of the $i$-th DFA module from bottom to top.

\section{Experiments}

\subsection{Datasets}
To evaluate MPCGNet, five publicly available polyp segmentation datasets are employed: CVC-ClinicDB~\cite{b6}, KvasirSEG~\cite{b7}, ETISETIS-LaribPolypDB~\cite{b8}, CVC-ColonDB~\cite{b9} and EndoScene~\cite{b10}. The dataset partitioning approach aligns with that of the networks used for comparison. A total of 1,450 images are used for training, comprising 900 images from Kvasir-SEG and 550 from CVC-ClinicDB. The leftover images from these two datasets serve as visible test sets to evaluate the network’s learning ability. The remaining three datasets are used as unseen test sets to measure the model’s generalization performance.

The performance of MPCGNet is evaluated using six metrics: mean Dice (mDice)~\cite{b28}, mean IoU (mIoU), the weighted F-measure ($\mathrm{F}_\mathrm{\beta}^\mathrm{w}$)~\cite{b29}, the structure measure ($\mathrm{S}_\mathrm{\alpha}$)~\cite{b30}, E-measure ($\mathrm{E}_\mathrm{\xi}$)~\cite{b31}, and mean absolute error (MAE).
\subsection{Implementation Details}
MPCGNet is developed using the PyTorch framework, trained and tested on an NVIDIA A800 GPU. All input images are uniformly resized to 352 × 352 pixels to ensure consistency. The training process spans 150 epochs with a batch size of 8, utilizing the AdamW optimizer initialized at a learning rate of 1e–4. The learning rate is halved every 50 epochs during training. To improve generalization and robustness, various data augmentation techniques—such as random rotation, random flipping, and color jitter—are employed to diversify the dataset.

\begin{table*}
\caption{The quantitative results of MPCGNet compared to other networks, with the highest scores shown in \textbf{bold} and the second-highest \uline{underlined}. The values are expressed in percentages(\%).}\label{tab1}
\captionsetup{justification=raggedright, singlelinecheck=false}
\centering
\setlength{\tabcolsep}{4mm}{
\begin{tabular}{l|l|l|l|l|l|l|l|l} 
\hline
Datasets                   & Networks     & Year & mDice          & mIoU           & $F_\beta^w$    & $S_\alpha$     & $E_\xi$        & MAE         \\ 
\hline\hline
\multirow{7}{*}{ETIS}      & PraNet     & 2020 & 76.54          & 68.38          & 69.88          & 86.05          & 87.99          & 1.81           \\
                           & HSNet      & 2022 & 80.12          & 71.51          & 75.20          & 87.99          & 91.74          & 1.38           \\
                           & CASCADE    & 2023 & 81.99          & \uline{74.90}  & 78.16          & 89.16          & \uline{92.34}  & 1.47           \\
                           & ECTransNet & 2023 & 77.80          & 69.67          & 75.68          & 85.09          & 88.64          & 3.58           \\
                           & MEGANet    & 2024 & 75.01          & 67.25          & 70.43          & 85.15          & 88.11          & 1.89           \\
                           & LHONet     & 2024 & \uline{82.29}  & 74.31          & \uline{77.90}  & \uline{89.37}  & \uline{91.92}  & \uline{1.36}   \\ 
\cline{2-9}
                           & MPCGNet    & ours & \textbf{84.49} & \textbf{76.45} & \textbf{80.02} & \textbf{90.51} & \textbf{93.25} & \textbf{1.19}  \\ 
\hline
\multirow{7}{*}{Kvasir}    & PraNet     & 2020 & 90.81          & 85.99          & 89.17          & 91.91          & 95.47          & 2.68           \\
                           & HSNet      & 2022 & 91.84          & 86.42          & 90.43          & 92.61          & 96.22          & 2.51           \\
                           & CASCADE    & 2023 & 91.87          & 86.97          & 90.78          & \uline{92.67}  & 96.37          & 2.34           \\
                           & ECTransNet & 2023 & 90.83          & 85.58          & 89.49          & 91.73          & 95.58          & 2.72           \\
                           & MEGANet    & 2024 & 90.76          & 85.51          & 89.39          & 91.60          & 94.77          & 2.94           \\
                           & LHONet     & 2024 & \uline{92.13}  & \uline{87.36}  & \uline{91.21}  & 92.64          & \textbf{96.80} & \uline{2.27}   \\ 
\cline{2-9}
                           & MPCGNet    & ours & \textbf{92.72} & \textbf{88.14} & \textbf{91.24} & \textbf{93.06} & \uline{96.39}  & \textbf{2.23}  \\ 
\hline
\multirow{7}{*}{ClinicDB}  & PraNet     & 2020 & 91.90          & 87.04          & 90.18          & 94.50          & 97.42          & 0.90           \\
                           & HSNet      & 2022 & 91.67          & 86.09          & 90.80          & 94.51          & 97.66          & 0.94           \\
                           & CASCADE    & 2023 & 93.15          & 88.62          & 92.10          & 95.41          & 97.77          & 0.74           \\
                           & ECTransNet & 2023 & 93.20          & 88.50          & 91.80          & 94.89          & 97.88          & 0.73           \\
                           & MEGANet    & 2024 & 94.06          & 89.46          & 93.33          & 95.48          & 98.49          & 0.71           \\
                           & LHONet     & 2024 & \uline{94.46}  & \uline{90.15}  & \uline{93.98}  & \uline{95.61}  & \uline{98.70}  & \textbf{0.56}  \\ 
\cline{2-9}
                           & MPCGNet    & ours & \textbf{94.70} & \textbf{90.25} & \textbf{94.06} & \textbf{95.67} & \textbf{99.05} & \uline{0.64}   \\ 
\hline
\multirow{7}{*}{ColonDB}   & PraNet     & 2020 & 76.16          & 69.07          & 74.13          & 84.85          & 87.68          & 3.90           \\
                           & HSNet      & 2022 & 79.37          & 70.87          & 76.99          & 86.23          & 90.73          & 3.20           \\
                           & CASCADE    & 2023 & 80.95          & 73.19          & 78.97          & 86.75          & 90.90          & 3.28           \\
                           & ECTransNet & 2023 & 77.80          & 69.67          & 75.68          & 85.09          & 88.64          & 3.58           \\
                           & MEGANet    & 2024 & 77.97          & 70.73          & 76.11          & 85.40          & 88.91          & 3.44           \\
                           & LHONet     & 2024 & \uline{82.00}  & \uline{73.97}  & \uline{79.57}  & \uline{86.98}  & \uline{91.70}  & \textbf{2.79}  \\ 
\cline{2-9}
                           & MPCGNet    & ours & \textbf{82.68} & \textbf{74.92} & \textbf{80.63} & \textbf{87.61} & \textbf{92.16} & \uline{3.06}   \\ 
\hline
\multirow{7}{*}{EndoScene} & PraNet     & 2020 & 88.55          & 81.32          & 85.21          & 92.94          & 96.02          & 0.83           \\
                           & HSNet      & 2022 & 89.07          & 82.34          & 86.48          & 93.26          & 96.23          & 0.79           \\
                           & CASCADE    & 2023 & 87.69          & 80.69          & 84.67          & 92.56          & 94.93          & 0.96           \\
                           & ECTransNet & 2023 & 88.86          & 82.38          & 86.60          & 93.01          & 97.05          & 0.69           \\
                           & MEGANet    & 2024 & 88.94          & 81.88          & 86.32          & 92.74          & 96.18          & 0.79           \\
                           & LHONet     & 2024 & \textbf{90.79} & \textbf{84.01} & \textbf{88.44} & \uline{93.75}  & \textbf{97.59} & \uline{0.56}   \\ 
\cline{2-9}
                           & MPCGNet    & ours & \uline{90.51}  & \uline{83.79}  & \uline{88.02}  & \textbf{93.88} & \uline{97.30}  & \textbf{0.53}  \\
\hline
\end{tabular}}
\end{table*}

\begin{figure*}
\centering
\includegraphics[width=1\linewidth]{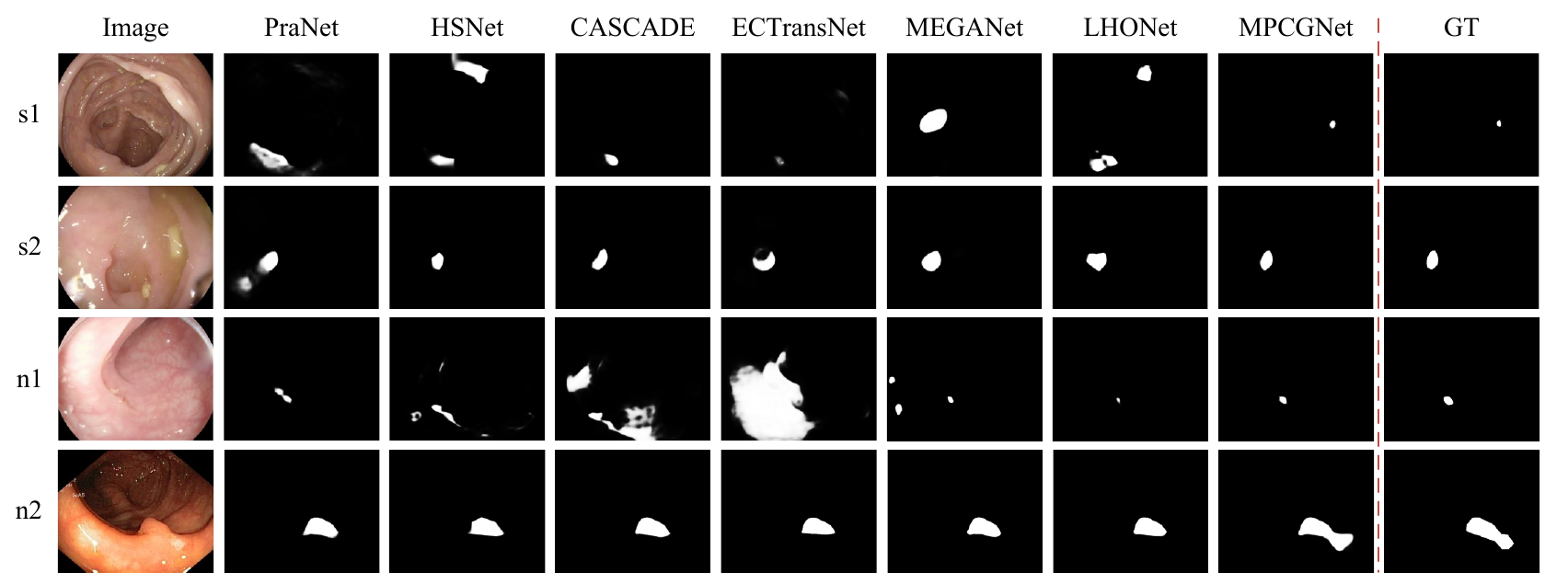}
\caption{The qualitative results of MPCGNet and other comparative networks.}
\label{compare}
\end{figure*}

\begin{table*}
\caption{The ablation experiment results of MPCGNet on EndoScene and ClinicDB. Coupling Gates (CG) indicates whether the CGMFE and DFA modules use coupling gates. If coupling gates are used, CG is marked with \checkmark.}\label{tab2}
\centering
\begin{tabular}{l|l|l|l|l|l|l|l|l|l} 
\hline
\multirow{2}{*}{CGMFE}      & \multirow{2}{*}{WCAD}       & \multirow{2}{*}{DFA}        & \multirow{2}{*}{CG}        & \multicolumn{3}{l|}{EndoScene} & \multicolumn{3}{l}{ClinicDB}  \\ 
\cline{5-10}
                            &                             &                             &                             & mDice & mIoU  & MAE            & mDice & mIoU  & MAE           \\ 
\hline\hline
\XSolidBrush & \Checkmark   & \Checkmark   & \Checkmark   & 89.58 & 82.65 & 0.74           & 92.88 & 88.40 & 0.71          \\ 
\hline
\Checkmark   & \XSolidBrush & \Checkmark   & \Checkmark   & 88.85 & 82.04 & 0.89           & 92.62 & 87.94 & 0.84          \\ 
\hline
\Checkmark   & \Checkmark   & \XSolidBrush & \Checkmark   & 88.92 & 82.43 & 0.88           & 93.75 & 89.04 & 0.67          \\ 
\hline
\Checkmark   & \Checkmark   & \Checkmark   & \XSolidBrush & 90.02 & 82.78 & 0.61           & 93.60 & 88.92 & 0.69          \\ 
\hline
\Checkmark   & \Checkmark   & \Checkmark   & \Checkmark   & \textbf{90.51} & \textbf{83.79} & \textbf{0.53}           & \textbf{94.70} & \textbf{90.25} & \textbf{0.64}          \\
\hline
\end{tabular}
\end{table*}

\begin{table}
\centering
\caption{The computational efficiency of MPCGNet and other comparison networks, where \textbf{bold} indicates the lowest values, and \uline{underlined} marks the second lowest. }\label{tab3}
\begin{tabular}{l|l|l|l|l|l} 
\hline
Networks   & Year & Backbone   & Params(M)     & FLOPs(G)      & FPS             \\ 
\hline\hline
PraNet     & 2020 & Res2Net-50 & 32.55          & 13.08          & \textbf{75.21}  \\ 
\hline
HSNet      & 2022 & PVTv2-B2   & \uline{29.85}  & \textbf{12.36} & \uline{57.23}   \\ 
\hline
CASCADE    & 2023 & PVTv2-B2   & 31.06          & 13.32          & 57.07           \\ 
\hline
ECTransNet & 2023 & Res2Net-50 & 51.99          & 81.60          & 30.47           \\ 
\hline
MEGANet    & 2024 & Res2Net-50 & 44.20          & 28.71          & 13.85           \\ 
\hline
LHONet     & 2024 & PVTv2-B2   & \textbf{28.06} & \uline{12.71}  & 24.41           \\ 
\hline
Ours       & Ours & PVTv2-B2   & 41.04          & 17.78          & 43.63           \\
\hline
\end{tabular}
\end{table}

\subsection{Results}
\subsubsection{Quantitative Results}
    MPCGNet is compared with six advanced networks, including PraNet~\cite{b11}, HSNet~\cite{b12}, CASCADE~\cite{b13}, ECTransNet~\cite{b14}, MEGANet~\cite{b15}, and LHONet~\cite{b16}. To guarantee a fair comparison, the code for each network was re-executed within our experimental environment. Owing to differences in experimental setups, the results may differ slightly from those reported in the original publications. The experimental results are presented in Table~\ref{tab1}. On the KvasirSEG and CVC-ClinicDB datasets, which are used to evaluate the learning ability of the models, MPCGNet achieves mDice scores that are 0.59\% and 0.24\% higher than the recently proposed LHONet, respectively, demonstrating a significant improvement in learning ability. MPCGNet also shows remarkable improvements in generalization ability on three other unseen datasets. The mDice scores on ETISETIS-LaribPolypDB and CVC-ColonDB surpass the second-best model by 2.20\% and 0.68\%, respectively. It is worth noting that while the mDice score of MPCGNet on EndoScene is slightly lower than that of LHONet, the overall generalization ability of MPCGNet demonstrates a significant improvement.

\subsubsection{Qualitative Results}
    The visualization results of MPCGNet compared to other networks are shown in Fig.~\ref{compare}. From images s1 and n1, it can be observed that MPCGNet achieves more accurate segmentation of small-sized colorectal polyps compared to other networks. In image s2, where the boundary between the left side of the polyp and the surrounding environment is unclear, MPCGNet accurately detects this boundary and successfully segments the polyp. Images n1 and n2 exhibit noise introduced by uneven lighting, resulting in inconsistent image brightness. Particularly in image n2, the presence of noise makes it challenging to detect certain parts of the polyp. However, MPCGNet leverages coupling gates to filter out noise, enabling the detection of these regions. Overall, MPCGNet demonstrates superior accuracy in detecting and segmenting polyps.
\subsubsection{Computational Efficiency}
    Table~\ref{tab3} presents the Params and FLOPs of MPCGNet in comparison with other networks. It can be observed that the Params and FLOPs of MPCGNet are moderate. Due to the use of an inverted bottleneck structure in the main modules, the Params and FLOPs are relatively higher compared to other lightweight networks. However, the advantages provided by the inverted bottleneck structure enable MPCGNet to achieve superior overall performance compared to other networks. Additionally, the inference speed of MPCGNet reaches 43.63 FPS, meeting the real-time requirements for polyp segmentation.

\subsection{Ablation Experiments}
To assess the contribution of each module and the coupling gates in MPCGNet, ablation studies were performed on the EndoScene and CVC-ClinicDB datasets. In each experiment, the CGMFE module, WCAD module, DFA module, and coupling gates were individually removed. The results, presented in Table~\ref{tab2}, show that the performance of MPCGNet declines regardless of which component is removed, indicating that each module is indispensable. When the WCAD module is removed, MPCGNet experiences the most significant performance drop, with mDice and mIoU decreasing by 1.66\% and 1.75\% on the EndoScene dataset, and by 2.08\% and 2.31\% on the ClinicDB dataset, respectively. This decline occurs because the WCAD module is essential for fusing deep semantic features with shallow features. Notably, coupling gates, which form a critical part of the CGMFE and DFA modules, play a significant role. When removed, the mDice and mIoU scores decrease by 0.49\% and 1.01\% on the EndoScene dataset and by 1.10\% and 1.33\% on the CVC-ClinicDB dataset, respectively. These results highlight the critical role of coupling gates in filtering noise and performing feature importance selection.

\section{Conclusion}
In this paper, A Multiscale Feature Extraction and Progressive Feature Aggregation Network Using Coupling Gates for Polyp Segmentation is proposed. The coupling gates multiscale feature extraction module aims to capture local features, filter noise, and bridge the encoder and decoder. The Windows cross attention decoder module is developed to fuse low-level features with high-level features. The decoder feature aggregation module is employed for large-scale feature aggregation and capture. The effectiveness of MPCGNet has been validated through experiments, demonstrating its ability to reduce misdiagnoses and missed diagnoses in colorectal polyp detection during colonoscopy, thereby providing patients with valuable time for timely treatment. However, MPCGNet shows limitations when handling cases with multiple polyps within a single image. This remains an area for potential improvement in future research.

\bibliography{MPCG}
\bibliographystyle{IEEEtran}

\end{document}